%% file: main.tex
\documentclass[10pt,twocolumn,letterpaper]{article}

\usepackage[pagenumbers]{cvpr}

\usepackage[T1,OT1]{fontenc}
\input{preamble}

\definecolor{cvprblue}{rgb}{0.21,0.49,0.74}
\usepackage[pagebackref,breaklinks,colorlinks,allcolors=cvprblue]{hyperref}

\def\confName{CVPR}
\def\confYear{2027}

\title{FewRef-5: Fixed-Reference Test-Time Adaptation for Virtual Staining}
\author{Wenxuan Wang$^{1}$\quad Liwen Wang$^{2}$\quad Mingmin Wu\quad Xingbo Dong$^{3}$\quad Zhe Jin$^{4}$\\[1pt]
\parbox{.97\textwidth}{\centering\fontsize{9}{10.5}\selectfont
Anhui Provincial Key Laboratory of Secure Artificial Intelligence, Anhui Provincial International Joint Research Center for Advanced Technology in Medical Imaging, School of Artificial Intelligence, Anhui University, Hefei 230093, China.}\\[1pt]
\parbox{.97\textwidth}{\centering\footnotesize
Email: $^{1}$\texttt{w126221302@stu.ahu.edu.cn}\quad $^{2}$\texttt{liwenwang@stu.ahu.edu.cn}\\
$^{3}$\texttt{xingbo.dong@ahu.edu.cn}\quad $^{4}$\texttt{jinzhe@ahu.edu.cn}}}

\hypersetup{pdfauthor={Wenxuan Wang, Liwen Wang, Mingmin Wu, Xingbo Dong, Zhe Jin},
pdftitle={FewRef-5: Fixed-Reference Test-Time Adaptation for Virtual Staining},
pdfsubject={Preprint}}
\makeatletter
\patchcmd{\@maketitle}{\vspace*{24pt}}{\vspace*{10pt}}{}{\PackageError{arxivcopy}{Title spacing patch failed}{}}
\patchcmd{\@maketitle}{\vspace*{12pt}}{\vspace*{6pt}}{}{\PackageError{arxivcopy}{Author spacing patch failed}{}}
\makeatother
\begin{document}
\lefthyphenmin=2
\righthyphenmin=3
\raggedbottom
\maketitle
\input{paper}
\paragraph{Tool-use disclosure.}
ChatGPT (OpenAI) assisted with manuscript drafting, translation and editing, and with analysis and figure-preparation code. The authors are responsible for the scientific content and reported results.
\clearpage
\twocolumn
{
    \raggedbottom
    \setlength{\bibsep}{1pt}
    \fontencoding{T1}\selectfont
    \small
    \bibliographystyle{ieeenat_fullname}
    \bibliography{main}
}
\clearpage
\onecolumn
\raggedbottom
\appendix
\input{appendix}

\end{document}

%% file: preamble.tex
\usepackage{currfile} 
\usepackage{array}
\usepackage{flafter}
\usepackage{stfloats}
\usepackage{caption} 

%% file: paper.tex
\begin{abstract}
Virtual staining synthesizes special-stain images from routine hematoxylin and eosin (H\&E), extending the information available from an existing digital slide. However, a fixed inference rule can leave available real target-stain references unused, limiting query-specific correction of stain appearance at deployment. Test-time adaptation (TTA) provides a mechanism for this reference-guided, query-specific refinement without retraining the generator. We therefore propose FewRef-5, a reference-assisted TTA method for virtual staining built on StainPromptInversion. We extract tissue-masked optical-density means and standard deviations from five real Masson's trichrome (MAS) training patches and average them into a fixed stain target. A scheduled statistic-matching loss introduces this target into per-query visual-prompt optimization alongside our edge/Gram consistency penalties and the inherited reconstruction objectives. The target remains unchanged across queries, avoiding recursive dependence on earlier generated outputs while the pretrained generator stays fixed. On recorded H\&E-to-MAS outputs for 1,989 queries, FewRef-5 reduces target-set FID from 183.201 to 166.904 and increases source-referenced SSIM from 0.92787 to 0.94996 relative to our Edge+Gram variant. Source PSNR and MS-SSIM decrease, showing an endpoint trade-off rather than uniformly better reconstruction.
\end{abstract}

\section{Introduction}
Virtual staining generates histological stain appearances from existing tissue images, including label-free measurements and routine hematoxylin and eosin (H\&E) slides~\cite{bai2023,rivenson2019auto,rivenson2019phase,dehaan2021}. In H\&E-to-special-stain translation, learned mappings and diffusion models synthesize target-domain appearance from an available input~\cite{isola2017,zhu2017,xiong2025}. Training establishes the synthesis relationship, but deployment raises another question: how can a particular prediction be revised using the current query and real target-stain references?

A fixed inference rule applies the learned mapping even when query appearance differs from training examples. Without an inference-time refinement step, available real target-stain references do not directly guide the adjustment of an individual prediction. Full-generator fine-tuning requires additional optimization and suitable supervision, while query-paired target images can be unavailable. This motivates an inference-time refinement mechanism that uses available real target-stain references to guide query-specific adaptation.

Test-time adaptation (TTA) uses inference-time information before delivering a prediction~\cite{sun2020,wang2021}. Visual prompts provide an editable interface for reference-guided refinement while generator weights remain fixed. StainPromptInversion already performs query-specific prompt optimization using source and style inversion trajectories~\cite{xiong2025}. We extend its objective with edge and Gram consistency losses and a fixed real-reference constraint. The test-time prompt-fitting interface is inherited from the base method.

The adaptation target matters. In our generated-memory variants, completed predictions supply statistics for later queries, making their correction signal depend on preceding outputs. Quality-gated writes select contributing predictions, but a reconstruction diagnostic does not independently certify target-stain appearance. Fixed real training references offer an alternative: remove recursive generated-output dependence from the stain target while fitting a prompt separately for each query.

We develop FewRef-5 as reference-assisted TTA on top of our edge/Gram-regularized StainPromptInversion pipeline. First, tissue-masked optical-density (OD) means and standard deviations from five real Masson's trichrome (MAS) training patches are averaged into a fixed target. Second, a scheduled OD-matching loss augments our edge/Gram-regularized reconstruction objective during prompt fitting. Third, the optimized prompt controls generation with fixed pretrained weights. No query-paired MAS enters adaptation, and generated outputs never update the reference statistic.

On recorded outputs for 1,989 H\&E-to-MAS queries, FewRef-5 reduces FID from 183.201 to 166.904 and increases source SSIM from 0.92787 to 0.94996 against our Edge+Gram variant. Source PSNR and MS-SSIM decrease. FID uses an unpaired real MAS set; the other endpoints use source H\&E. These retrospective comparisons expose a trade-off and do not establish generalization across reference sets. Our contributions are:
\begin{itemize}
\item We extend prompt adaptation with edge and Gram consistency losses and a scheduled OD constraint derived from fixed real references.
\item We establish reference-target order invariance and a fixed-state loss perturbation bound without asserting an image-quality guarantee.
\item We report full-set MAS endpoints and a verified same-input qualitative comparison, identifying metric references and the observed trade-offs.
\end{itemize}

\section{Related Work}
\subsection{Virtual Staining}
Virtual staining spans direct image translation, pathology-aware supervision, and diffusion-based synthesis~\cite{bai2023}. Paired adversarial mappings support virtual histology from label-free inputs and H\&E-to-special-stain translation~\cite{isola2017,rivenson2019auto,rivenson2019phase,dehaan2021,liu2022}; cycle consistency and patch contrastive learning enable unpaired mappings~\cite{zhu2017,park2020}. Stain normalization instead standardizes appearance within a stain~\cite{shaban2019,debel2019,gutierrez2022}. Pathology-aware methods use adaptive PatchNCE, protein/prototype constraints, weak supervision, or feature disentanglement to guide synthesis~\cite{li2023,chen2024psp,li2024weak,wang2025oda,xie2026}; benchmarking highlights the importance of data and validation protocols~\cite{klockner2025}. Diffusion provides iterative conditioning interfaces~\cite{song2021ddim,rombach2022}: PST-Diff uses pathological and structural constraints~\cite{he2024pst}, while StainPromptInversion fits query-specific visual prompts along inversion trajectories~\cite{xiong2025}. Our method adds edge/Gram regularization and a fixed five-reference OD constraint to this prompt interface. OD statistics themselves have earlier normalization precedents~\cite{macenko2009}.

\subsection{Test-Time Adaptation}
Relevant TTA methods use three overlapping interfaces. \textbf{Parameter adaptation} updates the predictor using self-supervision or entropy-related objectives~\cite{sun2020,wang2021,zhang2022memo}; CoTTA uses teacher/restoration mechanisms, and EATA/SAR control unreliable or unstable updates~\cite{wang2022cotta,niu2022eata,niu2023sar}. \textbf{Prompt adaptation} keeps the backbone fixed: TPT optimizes prompts across augmented views~\cite{shu2022tpt}, and VPTTA uses image-specific visual prompts with memory-assisted initialization~\cite{chen2024vptta}. TT-SaD adapts stain appearance for classification rather than special-stain synthesis~\cite{tsai2024ttsad}. \textbf{Memory/prototype adaptation} retains test information: T3A constructs pseudo-prototypes without backpropagation~\cite{iwasawa2021t3a}, while RoTTA manages balanced, timely, uncertainty-aware memory~\cite{yuan2023rotta}. FewRef-5 combines per-query prompt fitting with a fixed real-reference statistic; generated predictions do not update its target. Its reference-assisted setting is distinct from source-free adaptation.

\newpage
\begin{figure*}[t]
\centering
\includegraphics[width=\textwidth]{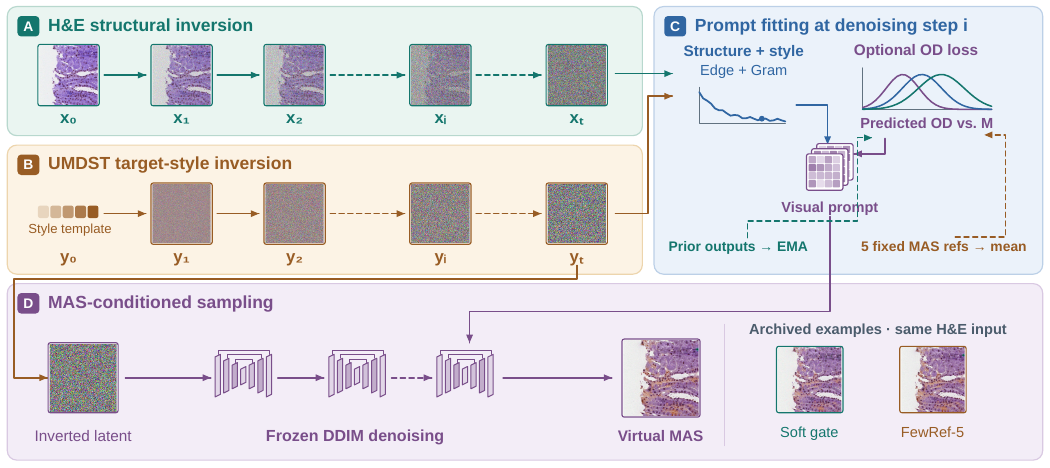}
\caption{Our adaptation built on StainPromptInversion: added edge/Gram consistency and the FewRef-5 constraint. Source and style inversions supply reconstruction targets. The prompt is fitted with fixed pretrained weights. Its optional OD term uses the frozen real-reference target in FewRef-5; the schematic symbol $M$ corresponds to $m_{\mathcal R}$. Intermediate states and curves are schematic. The MAS thumbnails are archived outputs for one shared input.}
\label{fig:method}
\end{figure*}

\section{Study Design and Adaptation Methods}
\begin{figure}[t]
\centering
\includegraphics[width=\columnwidth]{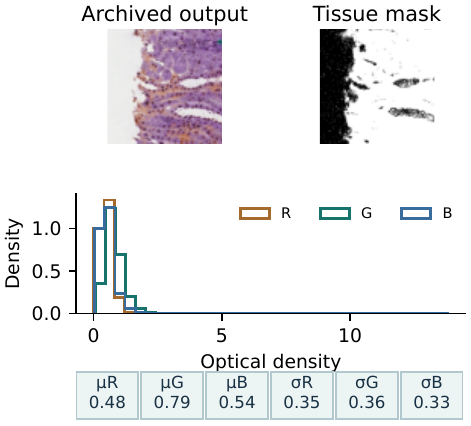}
\caption{Tissue masking and OD-statistic extraction illustrated on an archived FewRef-5 output. The binary mask and channel distributions are computed from that image. The reference target averages the same six statistics over five real training patches.}
\label{fig:statistics}
\end{figure}
\begin{figure}[t]
\centering
\includegraphics[width=\columnwidth]{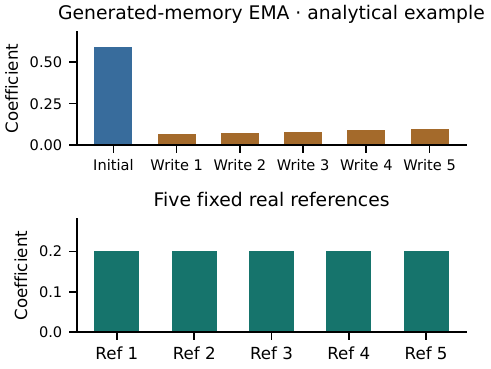}
\caption{Analytical coefficient illustration. For five EMA writes with $\beta=0.9$, the initialization retains weight $\beta^5$ and recent writes receive larger coefficients. The fixed reference average gives its five patches equal weight. These are formula-derived coefficients.}
\label{fig:weights}
\end{figure}

\subsection{General formulation and notation}
Let $x\in[-1,1]^{3\times H\times W}$ be source H\&E, $c_s,c_t$ source and target stain conditions, and $\theta$ fixed pretrained diffusion parameters. The GAN style adapter has fixed parameters $\psi$ and produces a template $a_\psi(x,c_t)$. A query-specific visual prompt is $\Phi=(\phi_{T-1},\ldots,\phi_0)$ with $\phi_i\in\mathbb R^{3\times H\times W}$. A set $\mathcal R=\{r_k\}_{k=1}^{K}$ contains real target-stain training patches. The corresponding real target image for $x$ is unavailable during prompt optimization.

A general inference-time objective is
\begin{equation}
\Phi^*(x;\mathcal R)\approx\arg\min_\Phi\sum_{i=0}^{T-1}\mathcal L_i(\Phi;x,\mathcal R).
\label{eq:general}
\end{equation}
The approximation denotes the implemented finite optimizer. The delivered image is $\hat y=G_\theta(x,c_t;\Phi^*)$, where $G_\theta$ includes the inversion initialization and prompted DDIM sampling. Table~\ref{tab:notation} defines the main symbols.

\subsection{Trajectories and our consistency losses}
Following StainPromptInversion~\cite{xiong2025}, inversion of $x$ supplies structural targets $V_i$, while inversion of $a_\psi(x,c_t)$ supplies style targets $U_i$. In the MAS setting, the source path is H\&E-conditioned and the style path is unconditional. At index $i$, a prompted deterministic step gives
\begin{equation}
(v_i,\hat y_i^0)=D_{\theta,i}(z_i+\phi_i;c_t),
\end{equation}
where $v_i$ is the next sample and $\hat y_i^0$ its predicted clean image. The style loss is $\mathcal L_{\rm sty}=\operatorname{MSE}(v_i,U_i)$. The local structural loss is
\begin{equation}
\mathcal L_{\rm str}=1-\frac{1}{|\Omega|}\sum_{p\in\Omega}
\frac{\sigma_{ab,p}+\epsilon_s}{\sqrt{\sigma^2_{a,p}\sigma^2_{b,p}}+\epsilon_s},
\end{equation}
with $a=\gamma(v_i)$, $b=\gamma(V_i)$, and $\gamma$ the replicated grayscale transform. Local moments use an $11\times11$ normalized Hann window; $\epsilon_s=10^{-8}$. This is the implementation's structural-correlation term. Appendix~\ref{app:implementation} identifies its numerical convention.

Our added penalties use $E(u)=|S_x*\gamma(u)|+|S_y*\gamma(u)|$ use grayscale Sobel responses, and let $F(u)$ flatten $\gamma(u)$ into $C\times HW$. Define
\begin{align}
\mathcal L_{\rm edge}&=\operatorname{MSE}(E(v_i),E(V_i)),\\
\Gamma(u)&=F(u)F(u)^\top/(CHW),\\
\mathcal L_{\rm Gram}&=\operatorname{MSE}(\Gamma(v_i),\Gamma(V_i)).
\end{align}
Our extended base objective is
\begin{align}
\mathcal L_{\rm base}={}&\alpha\mathcal L_{\rm sty}+(1-\alpha)\mathcal L_{\rm str}\notag\\
&+0.3\mathcal L_{\rm edge}+0.2\mathcal L_{\rm Gram}.
\label{eq:base}
\end{align}

\begin{figure*}[!t]
\centering
\includegraphics[width=0.92\textwidth]{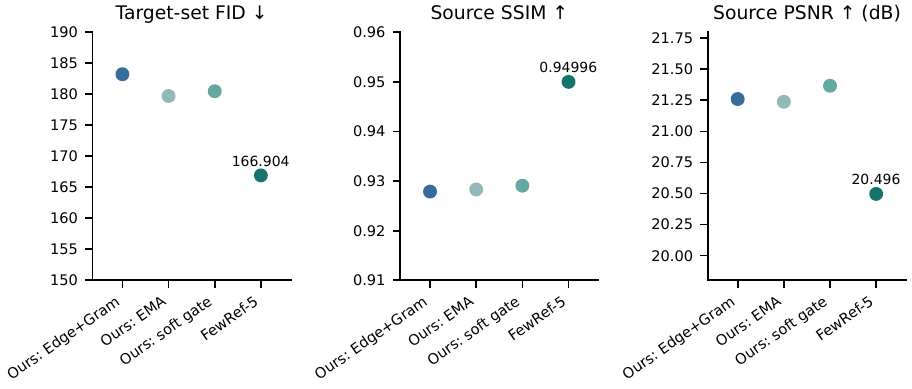}
\caption{Recorded MAS endpoints for selected local configurations. Target-set FID, source SSIM, and source PSNR have different references or sensitivities. Values reproduce Table~\ref{tab:mas}; repeated-run uncertainty is unavailable.}
\label{fig:endpoints}
\end{figure*}

\subsection{Tissue-aware optical-density target}
For $u\in[-1,1]^{3\times H\times W}$, convert each pixel to $I_{c,p}=\operatorname{clip}((u_{c,p}+1)/2,\epsilon,1)$ and $O_{c,p}=-\log I_{c,p}$, with $\epsilon=10^{-6}$. Define $b_p=\mathbf1[\sum_c I_{c,p}/3<0.90]$. The mask is $w_p=b_p$ if $\sum_pb_p\geq16$, and $w_p=1$ otherwise. For $N=\sum_pw_p$,
\begin{align}
\mu_c(u)&=\frac1N\sum_pw_pO_{c,p},\displaybreak[1]\\
\sigma_c(u)&=\sqrt{\frac1N\sum_pw_p(O_{c,p}-\mu_c)^2+\epsilon},\\
s(u)&=(\mu_R,\mu_G,\mu_B,\sigma_R,\sigma_G,\sigma_B).
\end{align}
\begin{table}[t]
\centering
\caption{Symbols and roles in fixed-reference prompt adaptation.}
\label{tab:notation}
\small
\begin{tabular}{@{}p{.20\columnwidth}p{.71\columnwidth}@{}}
\toprule
Symbol & Definition\\
\midrule
$x,c_s,c_t$ & H\&E query and source/target conditions\\
$\theta,\psi$ & Frozen diffusion and style-adapter parameters\\
$T,\Phi$ & DDIM steps and query-specific visual prompt\\
$\mathcal R,K$ & Real training reference set and its size\\
$s(u),m_{\mathcal R}$ & Six OD statistics and fixed reference average\\
$\alpha,\lambda$ & Style/structure balance and stain-loss weight\\
$\tau_i$ & Denoising-dependent stain-loss schedule\\
$U_i,V_i,z_i$ & Style target, structure target, current sample\\
\bottomrule
\end{tabular}
\end{table}

FewRef-5 computes statistics separately for each real training patch and then averages them,
\begin{equation}
m_{\mathcal R}=\frac1K\sum_{k=1}^{K}s(r_k),\qquad K=5.
\label{eq:reference}
\end{equation}
This vector stays fixed across query optimization and generation. Figure~\ref{fig:statistics} illustrates the statistic extraction. The method uses real target-domain references without query-paired target supervision.

\subsection{Scheduled prompt adaptation}
The stain constraint is $\ell_{\rm OD}(u;m)=\|s(u)-m\|_1/6$. At denoising index $i$,
\begin{align}
\tau_i&=1-\frac{i}{\max(T-1,1)},\\
\mathcal L_i&=\mathcal L_{\rm base}+\lambda\tau_i\ell_{\rm OD}(\hat y_i^0;m_{\mathcal R}).
\label{eq:full}
\end{align}
The increasing weight along denoising emphasizes cleaner predictions. For each query, prompts initialize at zero. At each index, Adam with learning rate $10^{-2}$ fits the prompt using $n_i=\lfloor50(1-i/T)\rfloor$ inner iterations. The sample is then advanced and the process continues to index zero. Final generation replays prompted DDIM steps from the style latent. The recorded MAS settings are $T=50$, $\alpha=0.05$, and $\lambda=0.01$. The update changes $\Phi$, while $\theta$, $\psi$, and $m_{\mathcal R}$ stay fixed.

\subsection{Properties of the fixed reference constraint}
\paragraph{Proposition 1.} The reference target in Eq.~\eqref{eq:reference} is invariant to reference-set ordering and query processing order. It has no recursive dependence on earlier generated outputs.

\begin{table*}[t]
\centering
\caption{Recorded H\&E-to-MAS endpoints. Local historical results use 1,989 queries: FID references unpaired real MAS, while the other metrics reference saved source H\&E. Header groups describe the local evaluation; published values follow the original protocol and are separated below. Bold marks the best recorded local value in each column.}
\label{tab:mas}
\small
\begin{tabular}{@{}lrrrrr@{}}
\toprule
Method / variant & Unpaired real MAS & \multicolumn{4}{c}{Source H\&E}\\
\cmidrule(lr){2-2}\cmidrule(l){3-6}
 & FID $\downarrow$ & PSNR (dB) $\uparrow$ & SSIM $\uparrow$ & CSS $\uparrow$ & MS-SSIM $\uparrow$\\
\midrule
Published StainPromptInversion~\cite{xiong2025} & 179.289 & 20.915 & 0.92330 & 0.93090 & 0.94580\\
\midrule
Ours (Edge+Gram) & 183.201 & 21.260 & 0.92787 & 0.93478 & \textbf{0.95486}\\
Ours (EMA memory) & 179.697 & 21.238 & 0.92830 & 0.93512 & 0.95386\\
Ours (hard gate 0.0035) & 182.180 & \textbf{21.630} & 0.93047 & 0.93659 & 0.95483\\
Ours (hard gate 0.0040) & 181.401 & 21.568 & 0.92992 & 0.93612 & 0.95438\\
Ours (soft gate) & 180.460 & 21.367 & 0.92906 & 0.93563 & 0.95422\\
FewRef-5 (ours) & \textbf{166.904} & 20.496 & \textbf{0.94996} & \textbf{0.95940} & 0.95071\\
\bottomrule
\end{tabular}
\end{table*}

\begin{figure*}[t]
\centering
\includegraphics[width=0.92\textwidth]{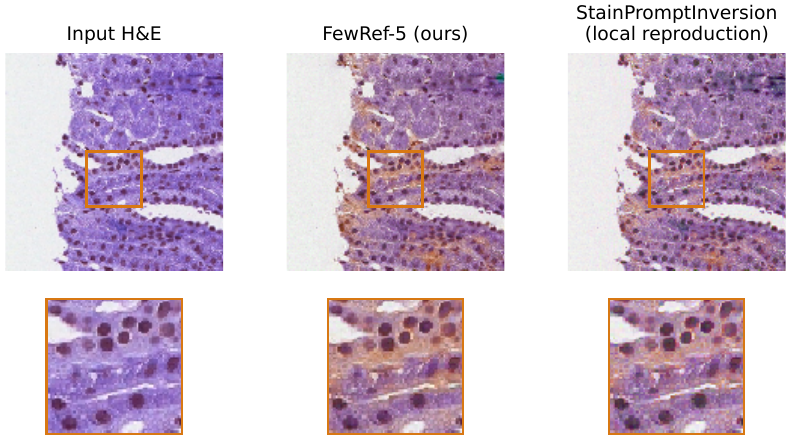}
\caption{Same-input H\&E-to-MAS comparison for \texttt{HE/0001.png}. Top: full patches; bottom: the same $64\times64$-pixel region (orange boxes), showing cellular detail, tissue--lumen boundaries, and local stain appearance. FewRef-5 (ours) adds edge/Gram and fixed-reference OD losses; the original-method reproduction uses only style/structure losses. The recorded checkpoint has a nominal 1,000,000-training-step label (previously ``1M''); inference uses 50 DDIM steps. No query-paired real MAS is available.}
\label{fig:comparison}
\end{figure*}

\paragraph{Proof.} A permutation of $\mathcal R$ preserves the finite sum. The expression contains only fixed training references and contains no query index or generated image. In comparison, an EMA memory with decay $\beta$ obeys
\begin{equation}
m_n=\beta^nm_0+(1-\beta)\sum_{j=1}^{n}\beta^{n-j}s(\hat y_j).
\end{equation}
Different temporal positions have different coefficients. Thus permutation invariance is a property of the fixed target; generated-memory updating has a different dependency structure. Figure~\ref{fig:weights} illustrates these coefficients.

\paragraph{Proposition 2.} For fixed image $u$ and two reference targets $m,m'$, the stain-loss difference satisfies
\begin{equation}
|\ell_{\rm OD}(u;m)-\ell_{\rm OD}(u;m')|\leq\frac{\|m-m'\|_1}{6}.
\end{equation}
If one reference statistic changes by $\Delta$, the full objective at fixed prompt, query, and trajectory targets changes by at most $\lambda\tau_i\|\Delta\|_1/(6K)$.

\paragraph{Proof.} Applying the reverse triangle inequality to $s(u)-m$ and $s(u)-m'$ gives the first bound. Replacing one summand in Eq.~\eqref{eq:reference} changes the average by $\Delta/K$. Multiplication by the nonnegative coefficient $\lambda\tau_i$ gives the second bound.

The propositions concern reference dependence and objective sensitivity. Different reference sets can have different biases. Global OD moments are also invariant to tissue-pixel permutations, leaving spatial assignment underdetermined. End-to-end image quality and optimization behavior require empirical evaluation.

\section{Evaluation and Recorded Results}
\subsection{Data and experimental setup}
\paragraph{Dataset and reference access.}
We evaluate the ANHIR-derived H\&E-to-MAS task on 1,989 RGB query patches at $256\times256$~\cite{borovec2020}. Five unique MAS training patches provide the fixed, equal-weight OD target; their identifiers remain in Appendix~\ref{app:implementation}. Optimization uses no query-paired MAS. FID uses an unpaired real MAS evaluation set; its exact size and patient/slide independence remain unverified.

\paragraph{Model and sampling configuration.}
The local launcher uses a pretrained diffusion model and a frozen GAN adapter. The diffusion model has base width 128, three residual blocks per level, class conditioning, linear noise scheduling, and rescaled timesteps. Batch size is one. Inference uses 50 deterministic DDIM steps, with unconditional style inversion, H\&E-conditioned structure inversion, and MAS-conditioned generation.

\paragraph{Execution environment.}
The completed same-input reproduction records an RTX 3090 (24 GB), Python 3.10.8, PyTorch 2.7.1+cu118, and CUDA runtime 11.8. These versions and recorded weight digests are verified for that reproduction. Complete environment snapshots and checkpoint binary equality across historical full-set runs remain unavailable.

\subsection{Adaptation settings and comparators}
\label{sec:adaptation-settings}
Prompts initialize at zero for each query. A fresh Adam optimizer at each denoising index uses learning rate $10^{-2}$ and the implemented integer conversion of $50(1-i/T)$ inner iterations. The recorded MAS weights are $\alpha=0.05$, $\lambda=0.01$, edge weight 0.3, and Gram weight 0.2. The OD schedule is $\tau_i=1-i/\max(T-1,1)$. The predicted clean image supplies the OD statistic, with a next-sample fallback in the implementation. Only prompts are optimized.

Table~\ref{tab:configuration-map} defines our method variants; Edge/Gram and the reference/memory constraints are our additions to the original style/structure objective. The gate diagnostic $q$ is the mean base loss at the final denoising index; it controls memory writes, not whether an output is delivered. All our generated-memory variants retain our Edge+Gram objective and scheduled OD term. These historical configurations jointly change reference provenance and update policy and are not a controlled one-factor ablation.

\begin{table}[!ht]
\centering
\caption{Configuration changes for the MAS comparison. Generated-memory rows use $\lambda=0.01$; published scores retain the original protocol.}
\label{tab:configuration-map}
\small
\begin{tabular}{@{}>{\raggedright\arraybackslash}p{.31\columnwidth}>{\raggedright\arraybackslash}p{.62\columnwidth}@{}}
\toprule
Method / variant & Objective or update rule\\
\midrule
Published & Original style/structure objective; external protocol.\\
Ours (Edge+Gram) & Adds $0.3\mathcal L_{\rm edge}+0.2\mathcal L_{\rm Gram}$; no OD-memory term.\\
Ours (EMA memory) & Adds OD loss; first output initializes memory, then EMA with $\beta=0.9$.\\
Ours (hard gates) & Write only for non-NaN $q\leq h$, with $h=0.0035$ or $0.0040$.\\
Ours (soft gate) & Write weight $0.1w(q)$, where $w(q)=\operatorname{clip}((0.0065-q)/0.003,0,1)$.\\
FewRef-5 (ours) & Matches the mean of five real training references; generated outputs never update it.\\
\bottomrule
\end{tabular}
\end{table}

\subsection{Evaluation protocol and record coverage}
FID compares generated outputs with the unpaired target-stain set~\cite{heusel2017}. PSNR, SSIM, contrast--structure similarity (CSS), and multiscale SSIM (MS-SSIM) compare each output with its saved source H\&E~\cite{wang2004,wang2003ms}. Thus the latter endpoints measure input similarity, not paired-target staining accuracy. We retain the archived FID implementation because preprocessing and estimation affect distributional scores~\cite{chong2020,parmar2022,binkowski2018}.

The analysis is retrospective and was not preregistered. EMA memory and soft-gate endpoints have archived evaluation logs; Edge+Gram and hard-gate values come from historical summaries; FewRef-5 values were supplied as terminal evaluation output. Published scores provide external protocol context. Exact ordered-query correspondence, repeated seeds, and reference resampling remain unverified. Appendix~\ref{app:provenance} retains the detailed source inventory.

\subsection{MAS results and their interpretation}
\paragraph{Generated-memory comparators.}
Table~\ref{tab:mas} shows that EMA memory lowers FID from 183.201 to 179.697 relative to our Edge+Gram, a decrease of 3.504. Hard gates at 0.0035 and 0.0040 give FIDs of 182.180 and 181.401, respectively; the soft gate gives 180.460. None improves FID over ungated EMA memory. The 0.0035 hard gate instead gives the largest recorded local PSNR, 21.630 dB. Increasing its threshold to 0.0040 lowers FID but also lowers PSNR and source SSIM. Thus the gate settings do not produce a common ranking across endpoints. The reconstruction-based write diagnostic does not independently measure agreement with real MAS.

\paragraph{Fixed real-reference adaptation.}
FewRef-5 reaches FID 166.904, below all listed local comparators and 12.793 below EMA memory. Against Edge+Gram, the decrease is 16.297 (8.90\%); source SSIM rises by 0.02209 and CSS by 0.02462. However, PSNR falls by 0.764 dB and MS-SSIM by 0.00415. Figure~\ref{fig:endpoints} visualizes this trade-off. FID evaluates an unpaired target-set distribution, whereas the other metrics reference source H\&E and use different aggregations. Their changes cannot be combined into a claim of uniformly better reconstruction or paired-target stain accuracy.

\paragraph{Interpretation of the comparison.}
The observed ranking is compatible with using real-reference statistics as a useful stain constraint in this recorded setting. It does not isolate reference provenance from the fixed update policy or quantify sensitivity to choosing another five patches. Global OD moments do not determine spatial stain placement. Moreover, these are single recorded comparisons without run-to-run intervals. The result therefore supports a descriptive endpoint improvement over the stated local comparators, while corresponding-region stain accuracy requires paired-target evidence.

\subsection{Same-input qualitative comparison}
Figure~\ref{fig:comparison} compares source H\&E, archived FewRef-5, and the completed original-method local reproduction for \texttt{HE/0001.png}. Pixel equality and input/output SHA-256 checks establish the shared processed input. The reproduction uses the original style/structure objective with edge, Gram, OD, and memory terms disabled; its verified record reports 1,268 inner iterations and a completed output. It supplies one qualitative query, not a full-set baseline. FewRef-5 additionally uses our edge/Gram penalties and fixed-reference OD loss. No query-paired real MAS is available; Appendix~\ref{app:qualitative} retains the digests and file-level provenance.

The shared query makes the visual comparison interpretable at the input level, but it does not isolate the OD term: the original-method reproduction also omits our edge and Gram penalties. Its one output cannot be used as a new full-set quantitative baseline. Consequently, the numerical gains above refer to Edge+Gram and the recorded memory comparators, not to a full-set rerun of the original two-term method.

\section{Discussion and Limitations}
The fixed reference target makes the stain constraint independent of earlier generated patches. It also introduces dependence on the chosen reference set. The loss perturbation bound concerns objective values at a fixed prompt, with optimizer trajectories and output fidelity outside that guarantee. The thresholded tissue mask makes the implemented objective piecewise defined. Finite Adam optimization supplies an approximate query-specific solution.

Full-set comparisons are retrospective and use one reference set. Some comparator values come from historical summaries, and FewRef-5 endpoints come from supplied evaluation output. Patient/slide independence, checkpoint identity, and ordered query correspondence remain incompletely documented. Appendix~\ref{app:pas} gives same-pipeline step/loss context; no FewRef-5 PAS endpoint is asserted. These limits constrain generalization and attribution.

\section{Conclusion}
We presented FewRef-5, a reference-assisted test-time adaptation method that augments our edge/Gram-regularized visual-prompt optimization with a fixed target derived from five real MAS training patches. Tissue-masked OD statistics provide an explicit stain constraint while the pretrained generator remains unchanged. The target has no recursive dependence on earlier generated images. Reference-order invariance and a fixed-state loss perturbation bound characterize this constraint without guaranteeing the final image quality.

On recorded outputs for 1,989 H\&E-to-MAS queries, FewRef-5 achieves lower FID and higher source-referenced SSIM and CSS than our Edge+Gram comparator, with decreases in PSNR and MS-SSIM. These results support the potential of real-reference guidance in this setting and reveal a trade-off among endpoints with different reference objects and sensitivities. The same-input illustration complements the quantitative records but supplies no paired-target validation.

Future work should assess localized staining fidelity using paired evidence, repeat experiments across reference selections and random seeds, and verify patient/slide separation. Comparisons under matched checkpoints and evaluation inputs would further clarify how reference provenance and update policy contribute to the observed differences.

%% file: appendix.tex
\section{Implementation and source evidence}
\label{app:implementation}
The method equations transcribe our archived FewRef-5 implementation. Edge and Gram consistency are our extensions to the original style/structure objective; FewRef-5 adds the fixed real-reference OD target. The structural loss uses the mean of the displayed local-correlation expression. Its window comes from an outer product of \texttt{torch.hann\_window(11)}, replicated across channels and normalized by the sum of that replicated tensor. Source and generated tensors are converted to grayscale using the torchvision coefficients and replicated to three channels. The Sobel kernels use the standard $3\times3$ horizontal and vertical integer filters. Edge magnitude is the sum of absolute responses. Gram matrices are calculated from the replicated grayscale tensor with normalization $CHW$.

Main Section~\ref{sec:adaptation-settings} gives the prompt optimizer, schedule, and OD evaluation point. The trajectory advances with each updated prompt; final generation begins at the style-inverted sample. Pretrained parameters receive no optimizer update.

The five reference basenames recorded in \texttt{references\_5.txt} are \texttt{patch1\_47\_99}, \texttt{patch2\_39\_93}, \texttt{patch3\_24\_188}, \texttt{patch3\_26\_104}, and \texttt{patch4\_21\_74}. The script requires five unique existing $256\times256$ RGB files. Their statistics are averaged with equal per-image weight. Resume checks compare the reference list and reconstructed statistics with the saved state. Generated outputs leave that target fixed.

\subsection{Historical generated-memory comparators}
EMA memory uses initialized EMA decay $\beta=0.9$. Hard gates use final-timestep mean base loss $q$ to permit writes at 0.0035 or 0.0040. The soft-gate write weight is
\begin{equation}
w(q)=\begin{cases}
1,&q\leq0.0035,\\
(0.0065-q)/0.003,&0.0035<q<0.0065,\\
0,&q\geq0.0065.
\end{cases}
\end{equation}
A new sample receives EMA weight $0.1w(q)$. An empty soft-gate memory initializes on its first full-weight sample. Comparing these memories with FewRef-5 changes both target provenance and update policy. The coefficient illustration in Figure~\ref{fig:weights} applies to an initialized memory and five subsequent writes.

\subsection{Result provenance}
\label{app:provenance}
Table~\ref{tab:provenance} identifies the available result sources. Checkpoint binary equality, exact ordered query lists, and repeated-run uncertainty are unverified. The three endpoint logs listed below are archived in their corresponding experiment folders. FewRef-5's final five values were supplied as terminal evaluation output after archive assembly. The complete evaluator transcript is unavailable in that archive. Source-code and log digests are listed in the accompanying ancillary file \texttt{anc/SOURCE\_RECORDS.tsv}.

\begin{center}
\begin{minipage}{\textwidth}
\centering
\captionof{table}{Source coverage for the retained pipeline. Endpoint-log coverage describes evaluation records, with full inference-log completeness unverified.}
\label{tab:provenance}
\small
\begin{tabular}{@{}p{.23\textwidth}p{.37\textwidth}p{.31\textwidth}@{}}
\toprule
Configuration & Source & Scope\\
\midrule
MAS Edge+Gram and hard gates; PAS Edge+Gram and two-term & Historical handoff summaries & Descriptive local endpoint records; complete evaluation logs unavailable\\
MAS EMA memory & Archived \texttt{eval\_1m.log} & Recorded full-set endpoints\\
MAS soft gate & Archived \texttt{sqg\_eval.log} & Recorded full-set endpoints\\
MAS FewRef-5 & Supplied terminal evaluation output & Five final endpoints; full evaluation transcript unavailable\\
PAS two-term, 100 steps & Archived \texttt{pas\_ds100\_eval.log} & Single recorded 100-step evaluation\\
MAS original-method reproduction & Supplied \texttt{run\_configuration.json} and \texttt{inference.log} & One qualitative query; no full-set endpoints\\
\bottomrule
\end{tabular}
\end{minipage}
\end{center}

\clearpage
\section{Additional same-pipeline PAS context}
\label{app:pas}
Our PAS variants compare Edge+Gram at 50 steps with the two-term objective at 50 and 100 steps. The two-term variants omit our edge/Gram penalties and use the style and source-structure terms with $\alpha=0.55$ and source conditioning in both inversion paths. The 50-to-100-step contrast changes step count under the two-term objective. A FewRef-5 PAS endpoint is unavailable. The local PAS FID target-set count also remains unverified.

\begin{center}
\begin{minipage}{\textwidth}
\centering
\captionof{table}{Recorded PAS context for the same base pipeline. The published row uses the original paper's protocol. Local FID compares against a target-stain set; other endpoints compare against source H\&E.}
\label{tab:pas}
\small
\begin{tabular}{@{}lcrrrrr@{}}
\toprule
Configuration & Steps & FID & PSNR & SSIM & CSS & MS-SSIM\\
\midrule
Published StainPromptInversion~\cite{xiong2025} & -- & 146.174 & 22.481 & 0.91140 & 0.91670 & 0.95000\\
Ours (Edge+Gram) & 50 & 156.703 & 23.302 & 0.87550 & 0.87900 & 0.96630\\
Ours (two-term) & 50 & 157.357 & 22.248 & 0.87320 & 0.87791 & 0.95100\\
Ours (two-term) & 100 & 156.544 & 22.587 & 0.90362 & 0.90855 & 0.95412\\
\bottomrule
\end{tabular}
\end{minipage}
\end{center}

\section{Qualitative comparison and provenance}
\label{app:qualitative}
Figure~\ref{fig:comparison} contains three verified same-input columns: H\&E, the archived FewRef-5 output, and the completed original-method local reproduction. The uploaded record confirms pixel equality for the processed query \texttt{HE/0001.png}. SHA-256 verification confirms the provided input and reproduced output match the completed run configuration. The code record matches the bundled original diffusion and structural-loss files. Edge, Gram, OD, and reference-memory terms are disabled in that reproduction. Each displayed zoom is the same unwarped $64\times64$ crop, with zero-based half-open bounds $[96,116,160,180)$ in $(x_0,y_0,x_1,y_1)$ order. The region was selected on the source H\&E, and the source image pixels are unchanged.

The reproduction uses local pretrained diffusion and GAN-adapter checkpoints. Their digests and runtime configuration appear in \texttt{anc/qualitative\_record/run\_configuration.json}; the settings and runtime are summarized in the main evaluation section. Both paths resolve to the local shared-weight directory. The former shorthand ``1M'' refers to the nominal 1,000,000-training-step label of the recorded checkpoint, not parameter count or adaptation iterations. Its training history is not independently verified by this inference record. Current digests do not establish equality with the historical FewRef-5 binaries. Native non-reentrant checkpointing supplies runtime compatibility.

The source package includes the completed configuration, source/code digests, preparation record, and inference log. The nested input manifest is the historical asset-collection record; its pending-output field predates this run. The top-level completed status and verified output digest describe the supplied result. The unpaired MAS examples have no established query correspondence and are excluded from the comparison. No query-paired ground truth is shown.

%% file: main.bbl
\begin{thebibliography}{40}
\providecommand{\natexlab}[1]{#1}
\providecommand{\url}[1]{\texttt{#1}}
\expandafter\ifx\csname urlstyle\endcsname\relax
  \providecommand{\doi}[1]{doi: #1}\else
  \providecommand{\doi}{doi: \begingroup \urlstyle{rm}\Url}\fi

\bibitem[Bai et~al.(2023)Bai, Yang, Li, Zhang, Pillar, and Ozcan]{bai2023}
Bijie Bai, Xilin Yang, Yuzhu Li, Yijie Zhang, Nir Pillar, and Aydogan Ozcan.
\newblock Deep learning-enabled virtual histological staining of biological
  samples.
\newblock \emph{Light: Science \& Applications}, 12:\penalty0 57, 2023.

\bibitem[Bi{\'n}kowski et~al.(2018)Bi{\'n}kowski, Sutherland, Arbel, and
  Gretton]{binkowski2018}
Miko{\l}aj Bi{\'n}kowski, Danica~J. Sutherland, Michael Arbel, and Arthur
  Gretton.
\newblock Demystifying {MMD GANs}.
\newblock In \emph{International Conference on Learning Representations}, 2018.

\bibitem[Borovec et~al.(2020)Borovec, Kybic, Arganda-Carreras, Sorokin,
  et~al.]{borovec2020}
Ji{\v r}{\'i} Borovec, Jan Kybic, Ignacio Arganda-Carreras, Dmitry~V. Sorokin,
  et~al.
\newblock {ANHIR}: Automatic non-rigid histological image registration
  challenge.
\newblock \emph{IEEE Transactions on Medical Imaging}, 39\penalty0
  (10):\penalty0 3042--3052, 2020.

\bibitem[Chen et~al.(2024{\natexlab{a}})Chen, Zhang, Zheng, Sun, He, and
  Qin]{chen2024psp}
Fuqiang Chen, Ranran Zhang, Boyun Zheng, Yiwen Sun, Jiahui He, and Wenjian Qin.
\newblock Pathological semantics-preserving learning for {H\&E}-to-{IHC}
  virtual staining.
\newblock In \emph{Medical Image Computing and Computer Assisted Intervention},
  2024{\natexlab{a}}.

\bibitem[Chen et~al.(2024{\natexlab{b}})Chen, Pan, Ye, Lu, and
  Xia]{chen2024vptta}
Ziyang Chen, Yongsheng Pan, Yiwen Ye, Mengkang Lu, and Yong Xia.
\newblock Each test image deserves a specific prompt: Continual test-time
  adaptation for 2d medical image segmentation.
\newblock In \emph{IEEE/CVF Conference on Computer Vision and Pattern
  Recognition}, pages 11184--11193, 2024{\natexlab{b}}.

\bibitem[Chong and Forsyth(2020)]{chong2020}
Min~Jin Chong and David Forsyth.
\newblock Effectively unbiased {FID} and inception score and where to find
  them.
\newblock In \emph{IEEE/CVF Conference on Computer Vision and Pattern
  Recognition}, 2020.

\bibitem[de~Bel et~al.(2019)de~Bel, Hermsen, Kers, van~der Laak, and
  Litjens]{debel2019}
Thomas de Bel, Meyke Hermsen, Jesper Kers, Jeroen van~der Laak, and Geert
  Litjens.
\newblock Stain-transforming cycle-consistent generative adversarial networks
  for improved segmentation of renal histopathology.
\newblock In \emph{Medical Imaging with Deep Learning}, pages 151--163, 2019.

\bibitem[de~Haan et~al.(2021)de~Haan, Zhang, Zuckerman, Liu, Sisk,
  et~al.]{dehaan2021}
Kevin de Haan, Yijie Zhang, Jonathan~E. Zuckerman, Tairan Liu, Anthony~E. Sisk,
  et~al.
\newblock Deep learning-based transformation of {H\&E} stained tissues into
  special stains.
\newblock \emph{Nature Communications}, 12:\penalty0 4884, 2021.

\bibitem[Guti{\'e}rrez~P{\'e}rez et~al.(2022)Guti{\'e}rrez~P{\'e}rez,
  Otero~Baguer, and Maass]{gutierrez2022}
Jos{\'e}~Carlos Guti{\'e}rrez~P{\'e}rez, Daniel Otero~Baguer, and Peter Maass.
\newblock {StainCUT}: Stain normalization with contrastive learning.
\newblock \emph{Journal of Imaging}, 8\penalty0 (7):\penalty0 202, 2022.

\bibitem[He et~al.(2024)He, Liu, Qi, Ding, Zhang, Song, Ma, Wu, Cai, Feng,
  Zhang, Zhang, and Zhang]{he2024pst}
Yufang He, Zeyu Liu, Mingxin Qi, Shengwei Ding, Peng Zhang, Fan Song, Chenbin
  Ma, Huijie Wu, Ruxin Cai, Youdan Feng, Haonan Zhang, Tianyi Zhang, and
  Guanglei Zhang.
\newblock {PST-Diff}: Achieving high-consistency stain transfer by diffusion
  models with pathological and structural constraints.
\newblock \emph{IEEE Transactions on Medical Imaging}, 43\penalty0
  (10):\penalty0 3634--3647, 2024.

\bibitem[Heusel et~al.(2017)Heusel, Ramsauer, Unterthiner, Nessler, and
  Hochreiter]{heusel2017}
Martin Heusel, Hubert Ramsauer, Thomas Unterthiner, Bernhard Nessler, and Sepp
  Hochreiter.
\newblock {GANs} trained by a two time-scale update rule converge to a local
  {Nash} equilibrium.
\newblock In \emph{Advances in Neural Information Processing Systems}, 2017.

\bibitem[Isola et~al.(2017)Isola, Zhu, Zhou, and Efros]{isola2017}
Phillip Isola, Jun-Yan Zhu, Tinghui Zhou, and Alexei~A. Efros.
\newblock Image-to-image translation with conditional adversarial networks.
\newblock In \emph{IEEE Conference on Computer Vision and Pattern Recognition},
  2017.

\bibitem[Iwasawa and Matsuo(2021)]{iwasawa2021t3a}
Yusuke Iwasawa and Yutaka Matsuo.
\newblock Test-time classifier adjustment module for model-agnostic domain
  generalization.
\newblock In \emph{Advances in Neural Information Processing Systems}, 2021.

\bibitem[Kl{\"o}ckner et~al.(2025)Kl{\"o}ckner, Teixeira, Montezuma, Fraga,
  Horlings, Cardoso, and Oliveira]{klockner2025}
Pascal Kl{\"o}ckner, Jos{\'e} Teixeira, Diana Montezuma, Jo{\~a}o Fraga,
  Hugo~M. Horlings, Jaime~S. Cardoso, and Sara~P. Oliveira.
\newblock {H\&E} to {IHC} virtual staining methods in breast cancer: An
  overview and benchmarking.
\newblock \emph{npj Digital Medicine}, 2025.

\bibitem[Li et~al.(2023)Li, Hu, Chen, and Kak]{li2023}
Fangda Li, Zhiqiang Hu, Wen Chen, and Avinash Kak.
\newblock Adaptive supervised {PatchNCE} loss for learning {H\&E}-to-{IHC}
  stain translation with inconsistent groundtruth image pairs.
\newblock In \emph{Medical Image Computing and Computer Assisted Intervention},
  2023.

\bibitem[Li et~al.(2024)Li, Dong, Huang, Li, Jiang, Fan, and Zhang]{li2024weak}
Jiahan Li, Jiuyang Dong, Shenjin Huang, Xi Li, Junjun Jiang, Xiaopeng Fan, and
  Yongbing Zhang.
\newblock Virtual immunohistochemistry staining for histological images
  assisted by weakly-supervised learning.
\newblock In \emph{IEEE/CVF Conference on Computer Vision and Pattern
  Recognition}, 2024.

\bibitem[Liu et~al.(2022)Liu, Zhu, Xu, Jia, Shi, and Jin]{liu2022}
Shengjie Liu, Chuang Zhu, Feng Xu, Xinyu Jia, Zhongyue Shi, and Mulan Jin.
\newblock {BCI}: Breast cancer immunohistochemical image generation through
  pyramid {Pix2pix}.
\newblock In \emph{IEEE/CVF Conference on Computer Vision and Pattern
  Recognition Workshops}, 2022.

\bibitem[Macenko et~al.(2009)Macenko, Niethammer, Marron, Borland, Woosley,
  Guan, Schmitt, and Thomas]{macenko2009}
Marc Macenko, Marc Niethammer, J.~S. Marron, David Borland, John~T. Woosley,
  Xiaojun Guan, Charles Schmitt, and Nancy~E. Thomas.
\newblock A method for normalizing histology slides for quantitative analysis.
\newblock In \emph{IEEE International Symposium on Biomedical Imaging}, 2009.

\bibitem[Niu et~al.(2022)Niu, Wu, Zhang, Chen, Zheng, Zhao, and
  Tan]{niu2022eata}
Shuaicheng Niu, Jiaxiang Wu, Yifan Zhang, Yaofo Chen, Shijian Zheng, Peilin
  Zhao, and Mingkui Tan.
\newblock Efficient test-time model adaptation without forgetting.
\newblock In \emph{International Conference on Machine Learning}, 2022.

\bibitem[Niu et~al.(2023)Niu, Wu, Zhang, Wen, Chen, Zhao, and Tan]{niu2023sar}
Shuaicheng Niu, Jiaxiang Wu, Yifan Zhang, Zhiquan Wen, Yaofo Chen, Peilin Zhao,
  and Mingkui Tan.
\newblock Towards stable test-time adaptation in dynamic wild world.
\newblock In \emph{International Conference on Learning Representations}, 2023.

\bibitem[Park et~al.(2020)Park, Efros, Zhang, and Zhu]{park2020}
Taesung Park, Alexei~A. Efros, Richard Zhang, and Jun-Yan Zhu.
\newblock Contrastive learning for unpaired image-to-image translation.
\newblock In \emph{European Conference on Computer Vision}, 2020.

\bibitem[Parmar et~al.(2022)Parmar, Zhang, and Zhu]{parmar2022}
Gaurav Parmar, Richard Zhang, and Jun-Yan Zhu.
\newblock On aliased resizing and surprising subtleties in {GAN} evaluation.
\newblock In \emph{IEEE/CVF Conference on Computer Vision and Pattern
  Recognition}, 2022.

\bibitem[Rivenson et~al.(2019{\natexlab{a}})Rivenson, de~Haan, Zhang, and
  Ozcan]{rivenson2019phase}
Yair Rivenson, Kevin de Haan, Yibo Zhang, and Aydogan Ozcan.
\newblock {PhaseStain}: The digital staining of label-free quantitative phase
  microscopy images using deep learning.
\newblock \emph{Light: Science \& Applications}, 8:\penalty0 23,
  2019{\natexlab{a}}.

\bibitem[Rivenson et~al.(2019{\natexlab{b}})Rivenson, Wang, Wei, de~Haan,
  Zhang, Wu, G{\"u}nayd{\i}n, Zuckerman, Chong, Sisk, Westbrook, Wallace, and
  Ozcan]{rivenson2019auto}
Yair Rivenson, Hongda Wang, Zhensong Wei, Kevin de Haan, Yibo Zhang, Yichen Wu,
  Harun G{\"u}nayd{\i}n, Jonathan~E. Zuckerman, Thomas Chong, Anthony~E. Sisk,
  Lindsey~M. Westbrook, W.~Dean Wallace, and Aydogan Ozcan.
\newblock Virtual histological staining of unlabelled tissue-autofluorescence
  images via deep learning.
\newblock \emph{Nature Biomedical Engineering}, 3:\penalty0 466--477,
  2019{\natexlab{b}}.

\bibitem[Rombach et~al.(2022)Rombach, Blattmann, Lorenz, Esser, and
  Ommer]{rombach2022}
Robin Rombach, Andreas Blattmann, Dominik Lorenz, Patrick Esser, and Bj{\"o}rn
  Ommer.
\newblock High-resolution image synthesis with latent diffusion models.
\newblock In \emph{IEEE/CVF Conference on Computer Vision and Pattern
  Recognition}, 2022.

\bibitem[Shaban et~al.(2019)Shaban, Baur, Navab, and Albarqouni]{shaban2019}
M.~Tarek Shaban, Christoph Baur, Nassir Navab, and Shadi Albarqouni.
\newblock {StainGAN}: Stain style transfer for digital histological images.
\newblock In \emph{IEEE International Symposium on Biomedical Imaging}, 2019.

\bibitem[Shu et~al.(2022)Shu, Nie, Huang, Yu, Goldstein, Anandkumar, and
  Xiao]{shu2022tpt}
Manli Shu, Weili Nie, De-An Huang, Zhiding Yu, Tom Goldstein, Anima Anandkumar,
  and Chaowei Xiao.
\newblock Test-time prompt tuning for zero-shot generalization in
  vision-language models.
\newblock In \emph{Advances in Neural Information Processing Systems}, 2022.

\bibitem[Song et~al.(2021)Song, Meng, and Ermon]{song2021ddim}
Jiaming Song, Chenlin Meng, and Stefano Ermon.
\newblock Denoising diffusion implicit models.
\newblock In \emph{International Conference on Learning Representations}, 2021.

\bibitem[Sun et~al.(2020)Sun, Wang, Liu, Miller, Efros, and Hardt]{sun2020}
Yu Sun, Xiaolong Wang, Zhuang Liu, John Miller, Alexei~A. Efros, and Moritz
  Hardt.
\newblock Test-time training with self-supervision for generalization under
  distribution shifts.
\newblock In \emph{International Conference on Machine Learning}, pages
  9229--9248, 2020.

\bibitem[Tsai et~al.(2024)Tsai, Chen, and Lu]{tsai2024ttsad}
Cheng-Chang Tsai, Yuan-Chih Chen, and Chun-Shien Lu.
\newblock Test-time stain adaptation with diffusion models for histopathology
  image classification.
\newblock In \emph{European Conference on Computer Vision}, pages 257--275,
  2024.

\bibitem[Wang et~al.(2021)Wang, Shelhamer, Liu, Olshausen, and
  Darrell]{wang2021}
Dequan Wang, Evan Shelhamer, Shaoteng Liu, Bruno Olshausen, and Trevor Darrell.
\newblock Tent: Fully test-time adaptation by entropy minimization.
\newblock In \emph{International Conference on Learning Representations}, 2021.

\bibitem[Wang et~al.(2022)Wang, Fink, Van~Gool, and Dai]{wang2022cotta}
Qin Wang, Olga Fink, Luc Van~Gool, and Dengxin Dai.
\newblock Continual test-time domain adaptation.
\newblock In \emph{IEEE/CVF Conference on Computer Vision and Pattern
  Recognition}, 2022.

\bibitem[Wang et~al.(2025)Wang, Wang, Wang, Wang, Xu, Cong, and
  Xu]{wang2025oda}
Tong Wang, Mingkang Wang, Zhongze Wang, Hongkai Wang, Qi Xu, Fengyu Cong, and
  Hongming Xu.
\newblock {ODA-GAN}: Orthogonal decoupling alignment {GAN} assisted by
  weakly-supervised learning for virtual immunohistochemistry staining.
\newblock In \emph{IEEE/CVF Conference on Computer Vision and Pattern
  Recognition}, 2025.

\bibitem[Wang et~al.(2003)Wang, Simoncelli, and Bovik]{wang2003ms}
Zhou Wang, Eero~P. Simoncelli, and Alan~C. Bovik.
\newblock Multiscale structural similarity for image quality assessment.
\newblock In \emph{Asilomar Conference on Signals, Systems and Computers},
  2003.

\bibitem[Wang et~al.(2004)Wang, Bovik, Sheikh, and Simoncelli]{wang2004}
Zhou Wang, Alan~C. Bovik, Hamid~R. Sheikh, and Eero~P. Simoncelli.
\newblock Image quality assessment: From error visibility to structural
  similarity.
\newblock \emph{IEEE Transactions on Image Processing}, 13\penalty0
  (4):\penalty0 600--612, 2004.

\bibitem[Xie et~al.(2026)Xie, Xu, Jiang, Rossi, K{\"a}rkk{\"a}inen, and
  Cong]{xie2026}
Shigeng Xie, Hongming Xu, Guiyang Jiang, Tuomo Rossi, Tommi K{\"a}rkk{\"a}inen,
  and Fengyu Cong.
\newblock Virtual immunohistochemistry staining with dual-aligned multi-task
  feature guidance.
\newblock In \emph{IEEE/CVF Conference on Computer Vision and Pattern
  Recognition}, 2026.

\bibitem[Xiong et~al.(2025)Xiong, Peng, Zhang, Chen, He, and Qin]{xiong2025}
Bing Xiong, Yue Peng, Ranran Zhang, Fuqiang Chen, Jiaye He, and Wenjian Qin.
\newblock Unpaired multi-domain histopathology virtual staining using dual path
  prompted inversion.
\newblock In \emph{Proceedings of the AAAI Conference on Artificial
  Intelligence}, pages 8780--8787, 2025.

\bibitem[Yuan et~al.(2023)Yuan, Xie, and Li]{yuan2023rotta}
Longhui Yuan, Binhui Xie, and Shuang Li.
\newblock Robust test-time adaptation in dynamic scenarios.
\newblock In \emph{IEEE/CVF Conference on Computer Vision and Pattern
  Recognition}, pages 15922--15932, 2023.

\bibitem[Zhang et~al.(2022)Zhang, Levine, and Finn]{zhang2022memo}
Marvin Zhang, Sergey Levine, and Chelsea Finn.
\newblock {MEMO}: Test time robustness via adaptation and augmentation.
\newblock In \emph{Advances in Neural Information Processing Systems}, 2022.

\bibitem[Zhu et~al.(2017)Zhu, Park, Isola, and Efros]{zhu2017}
Jun-Yan Zhu, Taesung Park, Phillip Isola, and Alexei~A. Efros.
\newblock Unpaired image-to-image translation using cycle-consistent
  adversarial networks.
\newblock In \emph{IEEE International Conference on Computer Vision}, 2017.

\end{thebibliography}
